\documentclass[times,onecolumn,preprint,a4paper]{elsarticle}	

\makeatletter
\def\ps@pprintTitle{%
 \let\@oddhead\@empty
 \let\@evenhead\@empty
 \def\@oddfoot{}%
 \let\@evenfoot\@oddfoot}
\makeatother

\usepackage{geometry}

\usepackage{framed,multirow}
\usepackage{booktabs}
\usepackage{tabu}
\usepackage{enumitem}
\newcommand*\rot{\rotatebox{90}}

\usepackage{amssymb}
\usepackage{latexsym}

\usepackage{amsmath}
\DeclareMathOperator{\Rot}{Rot}
\DeclareMathOperator{\Ref}{Ref}
\DeclareMathOperator{\Inv}{Inv}

\DeclareMathOperator*{\argmin}{arg\,min}

\usepackage{subcaption}

\usepackage{amsthm}
\theoremstyle{plain}
\newtheorem{theorem}{Theorem}[section]
\newtheorem{definition}{Definition}
\newtheorem{proposition}{Proposition}
\newtheorem{lemma}{Lemma}
\newtheorem{property}{Property}

\usepackage[hidelinks]{hyperref}
\usepackage[capitalise]{cleveref}
\newcommand{\figref}[1]{\figurename~\ref{#1}}
\newcommand{\etal}{\textit{et al.}}

\usepackage[dvipsnames]{xcolor}

\usepackage{algorithm}
\usepackage{fixltx2e}
\usepackage{algpseudocode}
\MakeRobust{\Call}
\algnewcommand\And{\textbf{and}}
\algnewcommand\Break{\textbf{break}}
\usepackage{varwidth}

\graphicspath{{images/},{control/}}

\usepackage{tikz}
\usepackage{tikzscale}

\newcommand{\experiment}[2]{
  \begin{minipage}[b]{.24\linewidth}
    \centering\includegraphics[width=.85\textwidth]{#1_01.png} %
    \subcaption{}\label{fig:#2a}
  \end{minipage}%
  \begin{minipage}[b]{.74\linewidth}
    \centering\includegraphics[width=.85\textwidth]{#1_03.pdf} %
    \subcaption{}\label{fig:#2b}
  \end{minipage}\\
  \begin{minipage}[b]{.24\linewidth}
    \centering\includegraphics[height=.85\textwidth]{#1_02.png} %
    \subcaption{}\label{fig:#2c}
  \end{minipage}%
  \begin{minipage}[b]{.74\linewidth}
    \centering\includegraphics[width=.85\textwidth]{#1_04.pdf} %
    \subcaption{}\label{fig:#2d}
  \end{minipage}
}

\begin{document}

\begin{frontmatter}

\title{Negentropic Planar Symmetry Detector}
\cortext[cor1]{Corresponding author: Department of Control Systems and Mechatronics, Faculty of Electronics, Wroc\l{}aw University of Technology, 50-370 Wroc\l{}aw, Poland}

\author[1]{A. Migalska\corref{cor1}} 
\ead{agata.migalska@gmail.com}
\author[2]{J.P. Lewis}

\address[1]{Wroc\l{}aw University of Technology, 50-370 Wroc\l{}aw, Poland}
\address[2]{Victoria University, Wellington 6012, New Zealand}

\begin{abstract}

In this paper we observe that information theoretical concepts are valuable tools for extracting information from images and, in particular, information on image symmetries. It is shown that the problem of detecting reflectional and rotational symmetries in a two-dimensional image can be reduced to the problem of detecting point-symmetry and periodicity in one-dimensional negentropy functions. Based on these findings a detector of reflectional and rotational global symmetries in greyscale images is constructed. We discuss the importance of high precision in symmetry detection in applications arising from quality control and illustrate how the proposed method satisfies this requirement. Finally, a superior performance of our method to other existing methods, demonstrated by the results of a rigorous experimental verification, is an indication that our approach rooted in information theory is a promising direction in a development of a robust and widely applicable symmetry detector. 

\end{abstract}

\end{frontmatter}

\section{Introduction}
Images are visual messages that convey information.  
A fundamental question is: what knowledge can we derive about an image while neither understanding its content nor inferring its meaning? The information theoretical approach to image and signal processing has been enjoying a renaissance, as its core concepts of entropy, mutual information and negentropy provide a precise and compact characterization of messages.
One phenomenon closely connected with information is symmetry. 
Intuitively, symmetry describes a situation in which there is information redundancy, with the same or similar information repeated in different regions related by the symmetry.
Therefore, the potential for applying information theoretic measures to symmetry detection is evident.

Symmetry detection is increasingly utilized in solving pressing issues in computer vision, such as image segmentation~\cite{gauch1993intensity,riklin2009symmetry}, shape representation and description~\cite{lee2013symmetry}, shape classification and recognition~\cite{harguess2011there}, image compression~\cite{sanchez2009symmetry}, image database indexing~\cite{sharvit1998symmetry}, structure recovery~\cite{pauly2008discovering,bokeloh2009symmetry}, and biometrics authentication~\cite{saber1998frontal,hayfron2003automatic}. There are several reasons for this increasing interest. First of all, symmetry is an omnipresent phenomenon, which makes symmetry detection applicable to multiple scenarios. Secondly, symmetry means redundancy and detecting it potentially reduces the amount of data and consequently simplifies further computations. Last but not least, this interest is inspired by the versatility of human visual processing in which symmetry is a highly salient feature~\cite{kootstra2008paying} as well as by biological significance of symmetry in humans and other animals \cite{dakin1998spatial,enquist1994symmetry}. 

\begin{figure}[htb]
\begin{minipage}[t]{.48\columnwidth}
\includegraphics[width=.9\textwidth]{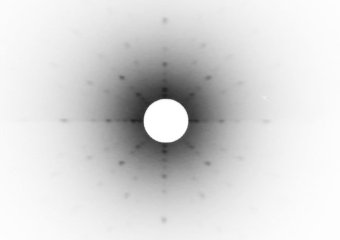}
\subcaption{}\label{fig:laue}
\end{minipage}
\begin{minipage}[t]{.48\columnwidth}
\includegraphics[width=.9\textwidth]{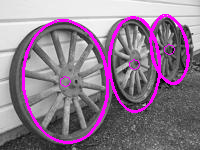}
\subcaption{}\label{fig:wheels}
\end{minipage}
\caption{\label{fig:laueVsWheels}\figref{fig:laue} -- Laue diffraction pattern from regular crystal \cite{laueimg}. Symmetry detection can be used to assert the lack of imperfections in a crystal. \figref{fig:wheels} - an image exhibiting local rotational symmetries under projection \cite{liu2013symmetry}. }
\end{figure}
A recent trend in symmetry detection is to focus on detecting approximate symmetries as well as  symmetries under projection or other deformations. For some applications, for instance for image segmentation, such an approach is not only acceptable but beneficial. However, there are applications in which high precision is critical and recognizing an approximate symmetry is not appropriate. 
Two examples of how the requirements for precision can vary are given in \figref{fig:laueVsWheels}.
One other application is quality control of manufactured items. When high precision is important it is of no surprise that appropriate attention is paid to the camera setup, which generally requires frontal photographs of an item under examination, and to image quality. Motivated by the importance of this application as well as by the insufficient precision of other approaches, we propose a new principle for symmetry detection. Upon that principle we develop a method for global symmetry detection. We demonstrate that our method outperforms several existing methods on the task of global rotational and reflectional symmetry detection in digital images.

\subsection{\label{sec:relatedWork}Related Work}
Methods for symmetry detection proposed to date comprise a voluminous catalogue and can be classified according to several key characteristics. The first characteristic is the type of symmetry. In this work we focus on methods capable of detecting both rotations and reflections. 

The second characteristic for classifying existing methods lies in whether symmetry is considered as an exact or as an approximate feature. 
The robustness of the exact approach in the context of digital images can be challenged on the grounds that discretization and sampling errors are enough to destroy any perfect symmetry. As well, it is believed that the human percept of symmetry accommodates a range of deviations from exact symmetry \cite{zabrodsky1994continuous}. In the light of these arguments, the work of Zabrodsky~\etal~\cite{zabrodsky1995symmetry} was a turning point in thinking about symmetry detection in images. This work introduced a continuous formulation of the problem capable of handling approximate symmetries. In what follows we only consider those methods that either do or can quantify symmetry on a continuous scale. 

Another criterion is whether the objective is to detect global or local symmetries. For global symmetries an entire image area serves as a symmetry support while local symmetries are defined on one or more smaller image regions where each region exhibits certain symmetry transformation, possibly varying among the regions. Local approaches to symmetry detection commonly draw on local feature extraction methods with the extracted features matched into pairs based on their generating transform. Masuda~\etal~\cite{masuda1993detection} described a method of extracting symmetries by performing correlation with the rotated and reflected images. This method incurs high computational cost and memory requirements, since all possible transformations (reflections, rotations, translations) have to be tried.  Podolak~\etal~\cite{podolak2006planar} detect reflection symmetries using a Monte-Carlo algorithm that selects a pair of surface points and votes for the plane between them. Mitra \etal \cite{mitra2006partial} cluster Hough-like votes for transformations that align boundaries with similar local shape descriptors. Loy and Eklundh \cite{loy2006detecting} quantify the ``amount'' of symmetry exhibited by each pair of features and the bilateral (mirror) and rotational symmetries are retrieved as top-voted ones in Hough-like voting space. 

Methods for global centred symmetry detection are commonly divided into two major categories. The first category is comprised of methods based on the Fourier transform, relying on the fact that symmetry of images is preserved in the frequency domain. Kazdhan~\etal~\cite{kazhdan2002reflective} proposed decomposing an image into concentric circles and computing the reflective symmetry descriptors on each of them to assess the level of reflectional symmetry. Derrode and Ghorbel~\cite{derrode2004shape} applied the Analytical Fourier-Mellin transform (AFTM) to detect rotational and reflectional symmetry, by solving for minimum the distance between the AFTM representations of two objects. Lucchese~\cite{lucchese2004frequency} presented an algorithm where the task of classifying symmetries is accomplished by looking at the point-wise zero crossings of the difference of the Fourier transform magnitudes along rays. Keller and Shkolnisky~\cite{keller2006signal} proposed an algorithm that is based on the properties of an angular correlation (AC), which is shown to be a periodic signal having frequency related to the order of symmetry. Chertok and Keller~\cite{chertok2010spectral} formulated the detection of symmetry of a set of points as the point alignment problem and proposed to solve it by means of spectral relaxation. 

The second category of methods for global symmetry detection is comprised of methods based on moments and moment invariants that satisfy certain constraints should an image exhibit a certain type of symmetry. 
Shen~\etal~\cite{shen1999symmetry} presented the Shen-Ip Symmetry Detector based on first three non-zero generalized complex (GC) moments. The properties of these three moments, that is their orders and repetitions, provide an information rich enough to retrieve the order and reflection axes of symmetries. 
Bissants~\etal~\cite{bissantz2009testing} proposed a framework of statistical tests for symmetries. Test statistics express certain restrictions on Zernike moments which are fulfilled should an image be symmetric. Their work was extended by Pawlak~\cite{pawlak2014statistical} with an estimator of the tilt angle of a reflection axis. 

In this paper we introduce a new approach to symmetry detection rooted in information theory.
Our approach is based on applying certain transformations to an input image and averaging the resulting image with the original one. A transformation can either be a reflection over an axis or a rotation by an angle. A transformed copy of the image is then ``put on top'' of the original image and the intensities of the two are averaged. When the image is averaged with its symmetrically transformed copy, the amount of information measured by entropy and, indirectly, by negentropy remains unchanged. Thus, the symmetry transform can be determined by comparing the negentropy of the original image with that of the image averaged with its copies under various candidate transformations.
Further, we demonstrate how the dimensionality of the problem can be reduced to a one-dimensional problems of periodicity and point-symmetry identification.
The work presented in this paper is a reformulation of initial work that we presented in \cite{migalska2015information}, where it was shown that an information theoretic principle can accurately guide retrieval of the axes of global reflectional symmetry in greyscale images. 

\subsection{Our Contribution}

Our contribution is four-fold. 
\begin{enumerate}
\item First, we introduce an information-theoretic formulation of image symmetry resulting in a simple procedure to determine the symmetries present in an image. 
\item Our second contribution is an evaluation of accuracy that our method and several other methods for symmetry detection achieve when confronted with the problem of global symmetry detection in greyscale images. We demonstrate that our method outperforms several state-of-the-art methods.
\item Our third contribution lies in an application of symmetry detection to a task arising in the domain of an automated visual inspection, namely shape inspection. We demonstrate how symmetry detection can guide decision making in quality control. 
\item Lastly, our final contribution is a ground-truth dataset for the global centred symmetry detection that we create and make publicly available~\cite{migalska2016groundtruth}.
\end{enumerate}

The remainder of the paper is organized as follows. Section~\ref{sec:theory} reviews the definitions of reflectional and rotational symmetries as well as the information theoretic characterization of messages. In Section~\ref{sec:method} the key findings regarding dimensionality reduction in symmetry retrieval are presented along with a comprehensive description of the algorithm. The results of the method's experimental verification are provided in Section~\ref{sec:experiments} and its application to automated shape inspection is described in Section~\ref{sec:control}. Discussion and concluding remarks are given in Section~\ref{sec:conclusion}.

\section{\label{sec:theory}Theoretical Preliminaries}

Our core objective is to derive knowledge about the information conveyed in an image. Consider, therefore, a two-dimensional image $I$ recorded on a square, equally-spaced Cartesian grid of size $n\times n$ whose intensities are in $\left[0,1\right]$. 

It is assumed that the origin of the coordinate system $O$ is located in the centre of an image. Then, a rotation about the origin $O$ by an angle $\theta$, denoted as $\Rot\left(\theta\right)$, is represented as a matrix 
\begin{align}
\Rot\left(\theta\right) ={\begin{bmatrix}\cos \theta &-\sin \theta \\\sin \theta &\cos \theta \end{bmatrix}},
\label{eq:rotationMatrix}
\end{align}
and a reflection about a line which passes through the origin and makes an angle $\theta$ with the x-axis, denoted as $\Ref\left(\theta\right)$, is represented as a matrix
\begin{align}
\Ref\left(\theta\right)={\begin{bmatrix}\cos 2\theta &\sin 2\theta \\\sin 2\theta &-\cos 2\theta \end{bmatrix}}.
\label{eq:reflectionMatrix}
\end{align}

If an object is invariant to one or both of the above transformations it is called symmetric. 
\begin{definition}[Rotational symmetry]
\label{def:rotationalSymmetry}
A function $\psi\colon\mathbf{R}^2\rightarrow \mathbf{R}$ is rotationally symmetric of order $K$ around the origin if 
\begin{equation}
\psi\left(x,y\right) = \left(\Rot\left(\beta_k\right)\psi\right)\left(x,y\right),
\end{equation}
where $\beta_k = \frac{360^\circ k}{K}$, $k=0,\dots, K-1$, and $\Rot\left(\beta_k\right)$ is a rotation transformation given by \eqref{eq:rotationMatrix}. 
\end{definition}

\begin{definition}[Reflectional symmetry]
\label{def:reflectionalSymmetry}
A function $\psi\colon\mathbf{R}^2\rightarrow \mathbf{R}$ is reflectionally symmetric with respect to the line which passes through the origin and makes an angle $\theta_0$ with the x-axis if 
\begin{equation}
\psi\left(x,y\right) = \left(\Ref\left(\theta_0\right)\psi\right)\left(x,y\right)
\label{eq:reflectionalSymmetry}
\end{equation}
where  
$\Ref\left(\theta_0\right)$ is a reflection transformation given by \eqref{eq:reflectionMatrix}. A function $\psi$ has reflectional symmetry of order $K$ if there are $K$ angles $\theta_k = \theta_0 + \frac{k \cdot 360^\circ}{K}$ that satisfy \eqref{eq:reflectionalSymmetry}, $k=0,\dots, K-1$.
\end{definition}

\begin{figure} 
\centering
\begin{minipage}[t]{.3\columnwidth}
\centering
\includegraphics[width=\textwidth]{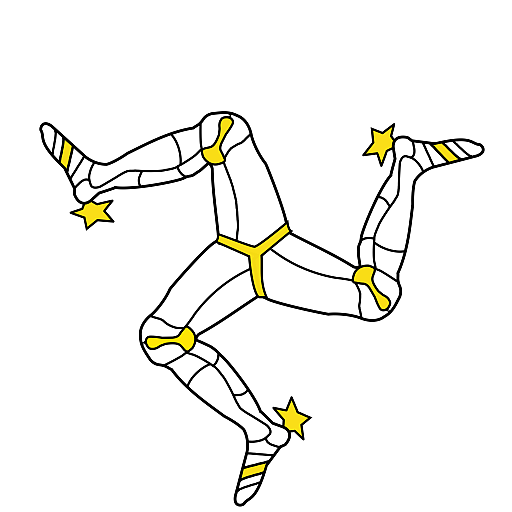}
\subcaption{\label{fig:rotational}}
\end{minipage}
\hfill
\begin{minipage}[t]{.3\columnwidth}
\centering
\includegraphics[width=\textwidth]{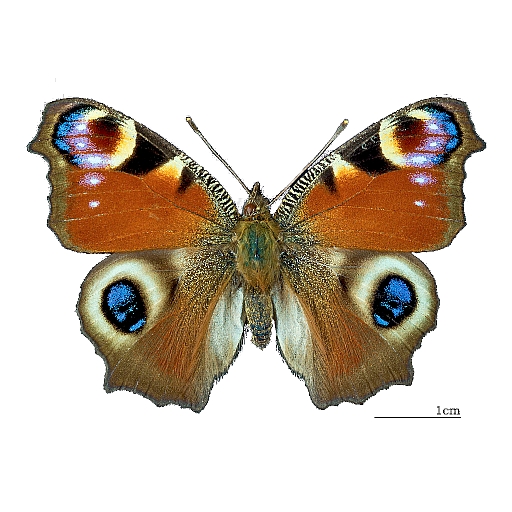}
\subcaption{\label{fig:mirror}}
\end{minipage}
\hfill
\begin{minipage}[t]{.3\columnwidth}
\centering
\includegraphics[width=\textwidth]{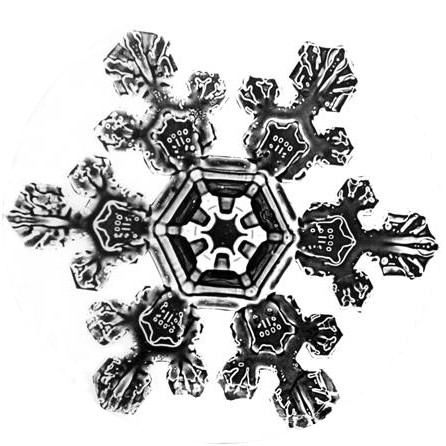}
\subcaption{\label{fig:reflectional}}
\end{minipage}
\caption{\label{fig:symmetric}Examples of symmetric images. \ref{fig:rotational} -The armoured triskelion on the flag of the Isle of Man~\cite{triskelion} is an example of rotational symmetry of order $3$. \ref{fig:mirror} - A butterfly has a reflectional symmetry of order $1$~\cite{paon}. \ref{fig:reflectional} - Every snowflake~\cite{snowflake} has a reflectional symmetry of order $6$ as well as a rotational symmetry of the same order.}
\end{figure}

Examples of symmetric images are given in \figref{fig:symmetric}.  

The following properties of object symmetries can be observed. 

\begin{property}
Every object has rotational symmetry of order $K=1$ as all objects are invariant under rotation by $\beta_1 = 360^\circ$.
\label{thm:property1}
\end{property}

It should be noted however that customarily an object invariant only to a rotation by $360^\circ / 1$ is not considered symmetric.

\begin{property}
Whenever an object has a reflectional symmetry of order $K$ then it also has a rotational symmetry of order $K$.
\label{thm:property2}
\end{property}

The proof of Property~\ref{thm:property2} is given in \ref{sec:appendix1}.

\subsection{The Amount of Information in an Averaged Message}

Our goal is to find all the symmetry transformations present in an image $I$. 
The symmetry determination will be done by examining the information content of the
image averaged with itself under various candidate transformations $T$
of the form \eqref{eq:rotationMatrix} or \eqref{eq:reflectionMatrix}, 
\begin{align}
I_T\left(i,j\right) = \frac{I\left(i,j\right) + \left(TI\right)\left(i,j\right)}{2},
\label{eq:summedRandVar}
\end{align}
where $i,j \in \{1,\dots,n\}$ and $T$ is the transformation used. 

As is common practice in the derivation of some image processing algorithms, we will regard images as continuous random signals. This viewpoint allows the application of differential entropy, an extension of Shannon entropy~\cite{shannon1949mathematical} to continuous probability distributions to measure the amount of information associated with a continuous random variable of pixel intensities. The differential entropy of a random variable $\mathbf{y}$ with probability density function (pdf) $p_y\left(\eta\right)$ is defined as
\begin{align}
H\left(\mathbf{y}\right) = -\int p_y\left(\eta\right) \log p_y\left(\eta\right) d\eta.
\end{align} 

In the case where the chosen transformation $T$ is the correct symmetry transform,
the entropy of the averaged image is equal to that of the original,
\[
   H\left( \frac{\mathbf{y} + T\mathbf{y}}{2} \right)
=   H\left( \frac{\mathbf{y} + \mathbf{y}}{2} \right)
=  H(\mathbf{y})
\]

On the other hand, if $T$ is incorrectly chosen, unrelated parts of the image will be summed,
and in the general case the entropy will change.
For example, summing unrelated parts of the image can be approximately regarded as summing independent random variables, which increases entropy if the variables are identically distributed,
and does so on average even without the i.i.d.~condition~\cite{artstein2004entropy}.

We remark that while other measures could be considered, the information theoretic
measures adopted here provide a simple but fundamental characterization of the problem.
For example, statistics such as the mean or variance could be considered.
However, the mean and variance are only particular moments of a random variable,
whereas entropy is a function of the full probability density and reflects all moments 
(the Fourier transform of a pdf is directly expressed through an infinite weighted sum of its moments).

Estimating entropy from the definition is computationally difficult, however it can be accurately approximated by means of the expected values of nonpolynomial functions \cite{hyvarinen1998new}:
\begin{align}
H\left(\mathbf{y}\right) &\approx H(\mathbf{z})  - \sum_{m=1}^M \xi_m^2 \Big(E\left\{G_m\left(\mathbf{y}\right)\right\} - E\left\{G_m\left(\mathbf{z}\right)\right\}\Big)^2 = H(\mathbf{z}) - J(\mathbf{y})
\label{eq:entropyApproximation}
\end{align}
where $E\{\cdot\}$ is an expected value, $\{G_m\}$ is a sequence of non-quadratic functions that do not grow too quickly, and $\xi_m^2$ are positive constants, scaling factors chosen so that functions $\{G_m\}$ form an orthonormal system and are orthogonal to all polynomials of second degree. 
The approximation \eqref{eq:entropyApproximation} utilizes negentropy $J\left(\mathbf{y}\right)$ which measures the difference between the amount of information in a given distribution and the normal distribution with the same mean and variance. 
We normalize the image intensities $\mathbf{y}$ to have zero mean and unit variance.
In this case $\mathbf{z}$ is a standard normal variable, of zero mean and unit variance, whose entropy $H(\mathbf{z})$ is equal to $\left(1 + \log\left(2\pi\right)\right)/2$. 

Instead of comparing the entropies of an original and of an averaged image, we argue that it is sufficient for computational purposes to compare their negentropies:
\begin{itemize}
\item After normalization, the negentropy differs from the entropy only by the sign and the constant offset $\left(1 + \log\left(2\pi\right)\right)/2$.
\item From the Central Limit Theorem, we know that the sum of random variables
tends toward a Gaussian distribution, and hence summing affects the negentropy.
\end{itemize}

In \cite{Hyvarinen2004} it is demonstrated that a good approximation of \eqref{eq:entropyApproximation} can be obtained with only two functions and is explicitly given by  
\begin{align}
J\left(\mathbf{y}\right) =& k_1 \bigg(E\left\{ \mathbf{y} \exp\left(-\mathbf{y}^2/2\right)\right\} \bigg)^2 + k_2 \left(E\left\{\exp\left(-\mathbf{y}^2/2\right)\right\} - \sqrt{\frac{1}{2}}\right)^2
\label{eq:negentropyApproximationFull}
\end{align}
where $k_1 = 36/\left(8\sqrt{3} - 9\right)$ and $k_2 = 24/\left(16\sqrt{3} - 27\right)$ \cite[pg. 119]{Hyvarinen2004}. 

\subsection{Application of Negentropy to Symmetry Detection}

We argue that the information on image symmetry can be determined from the information theoretical description of an image.  The following propositions are a direct consequence of the theoretical background presented up to here and form a basis for the construction of a symmetry detector that is blind to image content. 

\begin{proposition}
The order $K$ of symmetry is the maximal integer such that for each $k=1,\dots, K-1$
\begin{align}
J\left(I_{\Rot\left(k\cdot 360^\circ/K\right)}\right) = J\left(I\right).
\end{align}
\label{proposition1}
\end{proposition}

In other words, given an image averaged with its copy rotated by any multiple of the $K^{\textrm{th}}$ part of the whole angle, the order of symmetry is the maximal integer for which the negentropy of an averaged image is the same as the negentropy of an input image. 

\begin{proposition}
If an image $I$ has a reflectional symmetry of order $K$ with respect to $K$ lines rotated by $\theta_0 + k\cdot 180^\circ / K$ around the origin as well as a rotational symmetry of the same order $K$ then 
\begin{align}
J\left(I_{\Ref\left(\theta_0 + k\cdot 180^\circ / K\right)}\right) = J\left(I_{\Rot\left(l\cdot 360^\circ/K\right)}\right)
\end{align}
for any integer $k$ and $l$. 
\label{proposition2}
\end{proposition}

\section{\label{sec:method}Method}

In this section we present the key contribution of our paper -- the negentropic symmetry detector \textsc{NegSymmetry}. Given an input image $I$ \textsc{NegSymmetry} searches for two pieces of information -- an order of symmetry $K$ and the tilt angle of a reflection axis $\theta_0$. The method is derived from the following reasoning. Every object is invariant to at least one rotation, that is to a rotation by $360^\circ$, thus every object has a rotational symmetry of order $K\geq 1$.  In addition to a rotational symmetry, an object can also possess reflectional symmetry of the same order $K$, as per Property~\ref{thm:property2}. In order to determine if an image $I$ is invariant to any reflection the \textsc{NegSymmetry} algorithm searches for the tilt angle of a reflection axis. Its successful retrieval indicates that the image has reflectional symmetry, and an unsuccessful one indicates the existence of rotational symmetry only. Therefore these two pieces of information are enough to give a full description of planar symmetries existing in an image.

\subsection{\label{sec:step1}Step 1 - A Set of Candidate Orders of Symmetry}

Suppose the upper bound for the order of symmetry to be $K_{\max}$. For each $K$ between $2$ and $K_{\max}$ the image copy is obtained by rotating an original image by $360^\circ/K$. The original image is then averaged with the copy and the negentropy of a resulting image is calculated. Hereafter the negentropy values obtained in this way are referred to as \textit{rotational negentropy}. Formally, rotational negentropy, denoted as $J_{\Rot}$,  is a function from the set of natural orders of symmetry to the set of real-valued negentropies of an original image averaged with its copy obtained through the rotation by $360^\circ / K$, where $K$ is the order of symmetry,
\begin{align}
J_{\Rot}\left(I, K\right) = J\left(\frac{I + \Rot\left(360^\circ / K\right) I}{2}\right).
\label{eq:rotationalNegentropy}
\end{align}
Rotational negentropy for $K=1$, $J_{\Rot}\left(I,1\right)$, is equal to the negentropy of an original image and is referred to as a \textit{baseline negentropy}. 

An integer $K$ is considered eligible as the order of symmetry if the error between its rotational negentropy and the baseline negentropy does not differ more than a predefined threshold $\delta$, 
\begin{align}
\frac{\left|J_{\Rot}\left(I, K\right) - J_{\Rot}\left(I, 1\right)\right|}{J_{\Rot}\left(I, 1\right)} \leq \delta.
\label{eq:rotationSymCondition}
\end{align} 
In our experiments we have used $\delta=0.05$ and $\delta = 0.1$. From Equation~\eqref{eq:rotationSymCondition} one can obtain the tolerance bounds for $J_{\Rot}\left(I,k\right)$ given $\delta$,
\begin{align}
\left(1-\delta\right) \cdot J_{\Rot}\left(I,1\right) \leq J_{\Rot}\left(I,K\right) \leq \left(1+\delta\right) \cdot J_{\Rot}\left(I,1\right).
\label{eq:toleranceBounds}
\end{align}

The pseudocode of the symmetry order detection is presented in Algorithm~\ref{alg:rotationalSymmetry}.

\begin{algorithm}[htb]
\caption{\label{alg:rotationalSymmetry}RotationalNegentropy}
\begin{algorithmic}[1]
\Procedure {RotationalNegentropy}{$I$, $K_{\max}$}
\State $J \gets []$
\State $J[1] \gets$ \Call{Negentropy}{$I$} \Comment{Baseline negentropy}
\For{ $K \gets 2, K_{max}$} 
  \State $I_R \gets$ \Call{Rotate}{$I$, $360^\circ /K$}
  \State $I_A \gets (I + I_R)/2$
  \State $J[K] \gets$ \Call{Negentropy}{$I_A$}
\EndFor
\State \Return $J$
\EndProcedure
\end{algorithmic}
\end{algorithm} 

\subsection{\label{sec:step2}Step 2 - Verification of Periodicity}

Suppose an image copy is obtained by reflecting an original image across an axis rotated by some angle $\theta$. The negentropy of an average of the original image and the reflected copy is hereafter referred to as \textit{reflectional negentropy}. Reflectional negentropy, denoted as $J_{\Ref}$, is a function from the set of real-valued tilt angles to the set of real-valued negentropies of an original image averaged with the copy obtained through the reflection across the axis rotated by the tilt angle in question,
\begin{align}
J_{\Ref} \left(I, \theta\right) = J\left(\frac{I + \Ref\left(\theta\right) I}{2}\right).
\label{eq:reflectionalNegentropy}
\end{align}

The following theorem presents a favourable characteristic of reflectional negentropy that forms a basis for a decision rule selecting the order of symmetry.

\begin{theorem}
\label{thm:theorem2}
If an image has rotational symmetry of order $K$ then its reflectional negentropy is a periodic function with the period of $180^\circ/K$. 
\end{theorem}

\begin{proof}
A function $f$ is said to be periodic with period $P$ ($P$ being a non-zero constant) if 
\[
f\left(x+P\right) = f\left(x\right)
\]
for all values of $x$ in the domain. 

Let $I$ be an image that has rotational symmetry of order $K$ and let $\theta$ be an arbitrary angle. Consider $K$ lines $\{L_l\}$ rotated by $\theta + l\cdot 180^\circ/K$ about the origin, $l=0,\dots, K-1$, as well as $K$ copies of the image $I$ resulting from reflecting $I$ across $L_l$. Note that all $K$ copies are the same, thus the reflectional negentropy attains the same value for all $K$ averaged images. As the lines $\{L_l\}$ are rotated by the angles distant from one another by $180^\circ/K$, thus $J\left(\theta + 180^\circ/K\right) = J\left(\theta\right)$. 
\end{proof}

\begin{figure}[htb]
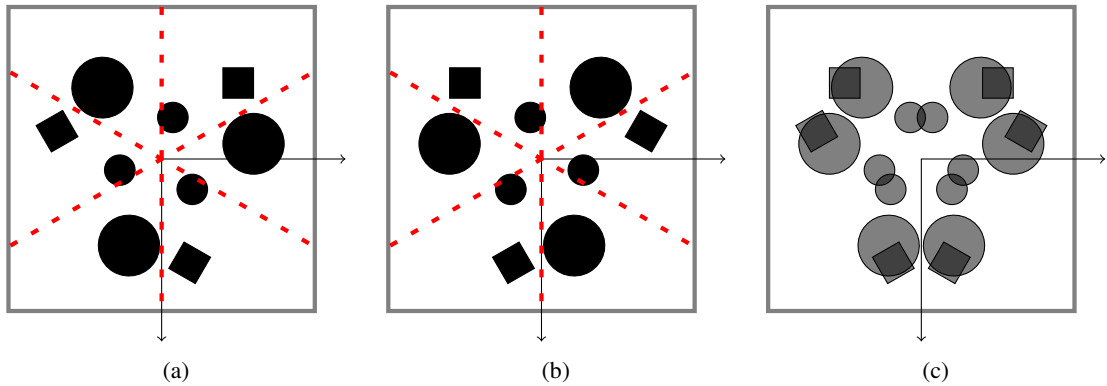

\begin{minipage}{.32\columnwidth}
\centering
\includegraphics[width=.95\textwidth]{th2_im1.tikz}
\subcaption{\label{fig:th2_im1}}
\end{minipage}\hfill
\begin{minipage}{.32\columnwidth}
\centering
\includegraphics[width=.95\textwidth]{th2_im2.tikz}
\subcaption{\label{fig:th2_im2}}
\end{minipage}\hfill
\begin{minipage}{.32\columnwidth}
\centering
\includegraphics[width=.95\textwidth]{th2_im3.tikz}
\subcaption{\label{fig:th2_im3}}
\end{minipage}
\caption{\label{fig:th2} Image \ref{fig:th2_im1} is rotationally symmetric with the order of symmetry $K=3$. Red dashed lines $L_i$ are the reflection axes rotated by $\theta + i\cdot 360^\circ/K = \theta +  i\cdot 120^\circ$ about the origin, $i=0, 1, 2$. No matter which of these three reflection axes is chosen the reflected copies look as in \ref{fig:th2_im2}. The original image \ref{fig:th2_im1} averaged with its copy \ref{fig:th2_im2} is presented in \ref{fig:th2_im3} -- the same averaged image is obtained for each $L_i$.}
\end{figure}

\begin{figure}[htb]
\centering
\includegraphics[width=.95\columnwidth]{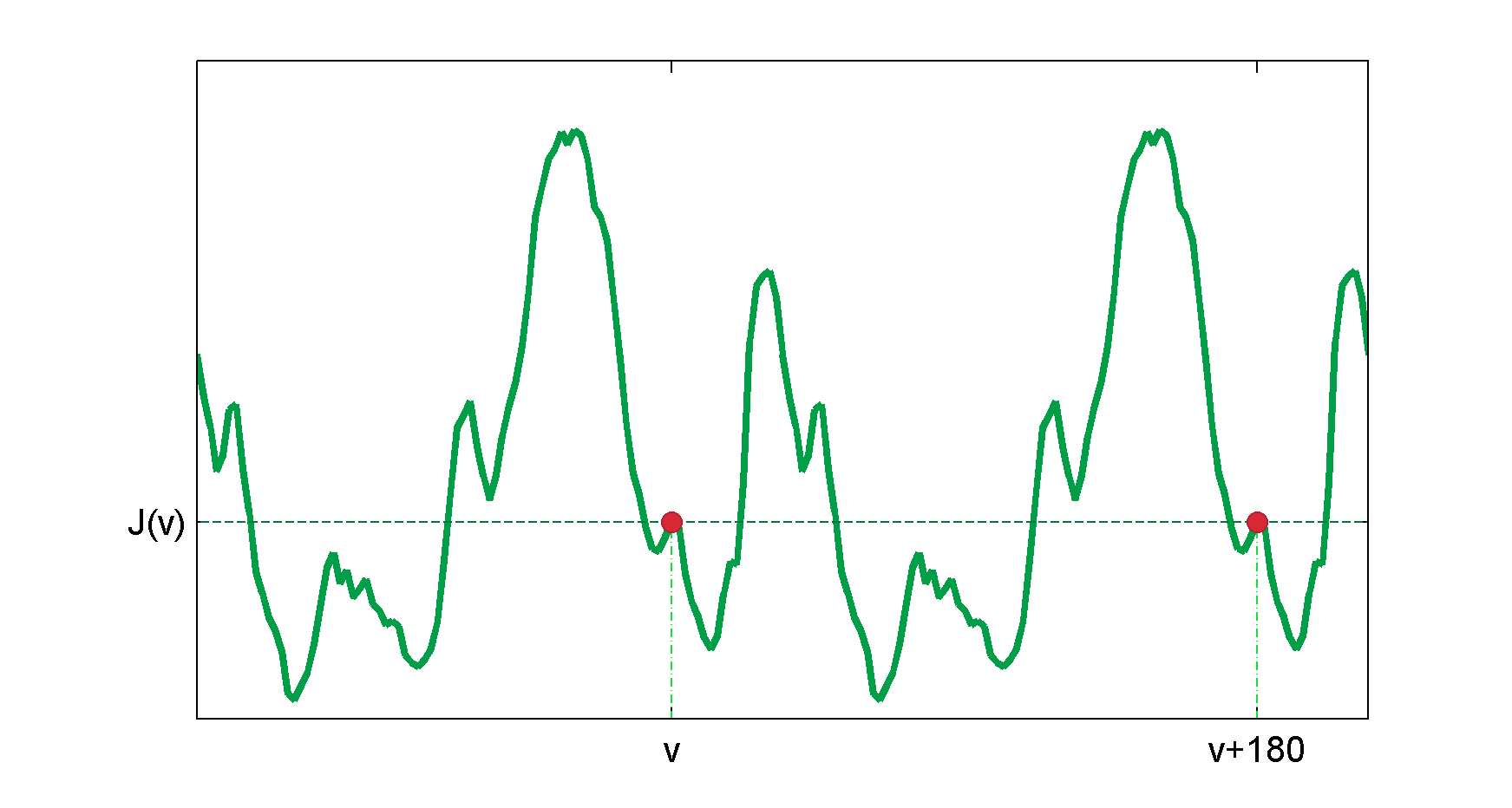}
\caption{\label{fig:periodicity}Reflectional negentropy of a symmetric image averaged with a transformed copy of itself is periodic with the period of $180^\circ / K$, where $K$ is the order of symmetry. The period of $180^\circ$ is attained if an image exhibits invariance only to rotation by the whole angle.}
\end{figure}

A further explanation of the above theorem is presented in \figref{fig:th2} and in \figref{fig:periodicity}.

Based on the above theorem we propose the following decision rule selecting the order of symmetry: 
\begin{proposition}
Given a vector of candidate orders of symmetry choose the largest one for which the reflectional negentropy satisfies the periodicity condition given in Theorem~\ref{thm:theorem2}.
\label{thm:proposition3}
\end{proposition}

\subsection{Step 3 - Retrieval of a Tilt Angle}
 
The remaining missing piece of information is the tilt angle of one of the reflection axes. 
\begin{theorem}
\label{thm:theorem1}
Let an image $I$ have reflectional symmetry across a line rotated by $\theta$ about the origin. Then the reflectional negentropy posses the point symmetry across the point $\theta$.
\begin{align}
I = \Ref\left(\theta\right) I \quad \Rightarrow \quad J_{\Ref}(I) = \Inv_\theta\left(J_{\Ref}\left(I\right)\right),
\end{align}
where $\Inv_\theta$ is a point symmetry across the point $\theta$. 
\end{theorem}
\begin{proof}
Given a one-dimensional real-valued vector $\mathbf{a}$ the formula for the point reflection $p$ (or inversion in the point $p$) is $\Inv_p\left(\mathbf{a}\right) = 2p - \mathbf{a}$. Equivalently, if $b=p-a$ then $a = p-b$ and $2p-a = p+b$, then $\Inv_p\left(p-b\right) = p+b$.

Let an image $I$ be reflectionally symmetric with respect to a line $L$ rotated by $\theta$ about the origin. Let $\phi$ be an arbitrary angle, $\phi \in [0, 180]$. 
Let $\Ref\left(\theta + \phi\right) I$ and $\Ref\left(\theta - \phi\right) I$ be the copies of $I$ obtained by reflecting $I$ across the lines tilted by $\theta+\phi$ and $\theta - \phi$, respectively. Due to the fact that $I$ is reflectionally symmetric with respect to the line $L$, the images $\Ref_{\theta + \phi} I$ and $\Ref_{\theta - \phi}I$ are also reflectionally symmetric to each other with respect to the line $L$. Consequently, the images $I_{\Ref\left(\theta - \phi\right)}$ and $I_{\Ref\left(\theta+\phi\right)}$, obtained by averaging the original image with each of the copies, are reflectionally symmetric to each other with respect to the line $L$. Thus, the negentropies of $I_{\Ref\left(\theta - \phi\right)}$ and $I_{\Ref\left(\theta+\phi\right)}$ are equal, $J\left(I_{\Ref\left(\theta - \phi\right)}\right) = J\left(I_{\Ref\left(\theta+\phi\right)}\right)$.
\end{proof}

\begin{figure}[htb]
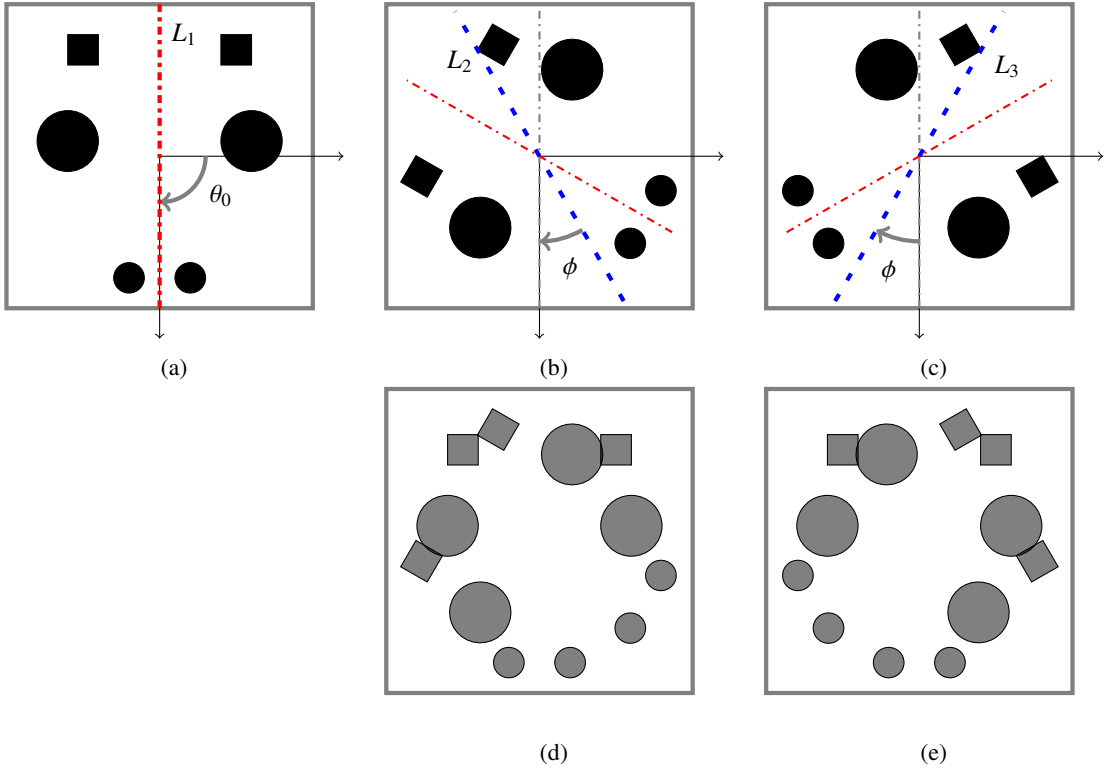

\begin{minipage}{.32\columnwidth}
\centering
\includegraphics[width=.95\textwidth]{th1_im1.tikz}
\subcaption{\label{fig:th1_im1}}
\end{minipage} \hfill
\begin{minipage}{.32\columnwidth}
\centering
\includegraphics[width=.95\textwidth]{th1_im2.tikz}
\subcaption{\label{fig:th1_im2}}
\end{minipage} \hfill
\begin{minipage}{.32\columnwidth}
\centering
\includegraphics[width=.95\textwidth]{th1_im3.tikz}
\subcaption{\label{fig:th1_im3}}
\end{minipage}\,
\begin{minipage}{.32\columnwidth}
\centering
\includegraphics[width=.95\textwidth]{th1_empty.tikz}
\end{minipage} \hfill
\begin{minipage}{.32\columnwidth}
\centering
\includegraphics[width=.95\textwidth]{th1_im4.tikz}
\subcaption{\label{fig:th1_im4}}
\end{minipage} \hfill
\begin{minipage}{.32\columnwidth}
\centering
\includegraphics[width=.95\textwidth]{th1_im5.tikz}
\subcaption{\label{fig:th1_im5}}
\end{minipage}
\caption{\label{fig:th1}Image \ref{fig:th1_im1} is reflectionally symmetric with respect to a dash-dotted red line $L_1$ rotated by $\theta_0$ about the origin of the coordinate system. The image is then reflected across a blue dashed line $L_2$ rotated by $\theta_0 - \phi$ (\ref{fig:th1_im2}), and across a blue dashed line $L_3$ rotated by $\theta_0 + \phi$ (\ref{fig:th1_im3}). Images \ref{fig:th1_im2} and \ref{fig:th1_im2} are symmetric to each other with respect to the line $L_1$ rotated by $\theta_0$. \ref{fig:th1_im4} - image \ref{fig:th1_im1} averaged with \ref{fig:th1_im2}; \ref{fig:th1_im5} - image \ref{fig:th1_im1} averaged with \ref{fig:th1_im3}.}
\end{figure}

\begin{figure}[htb]
\centering
\includegraphics[width=.95\columnwidth]{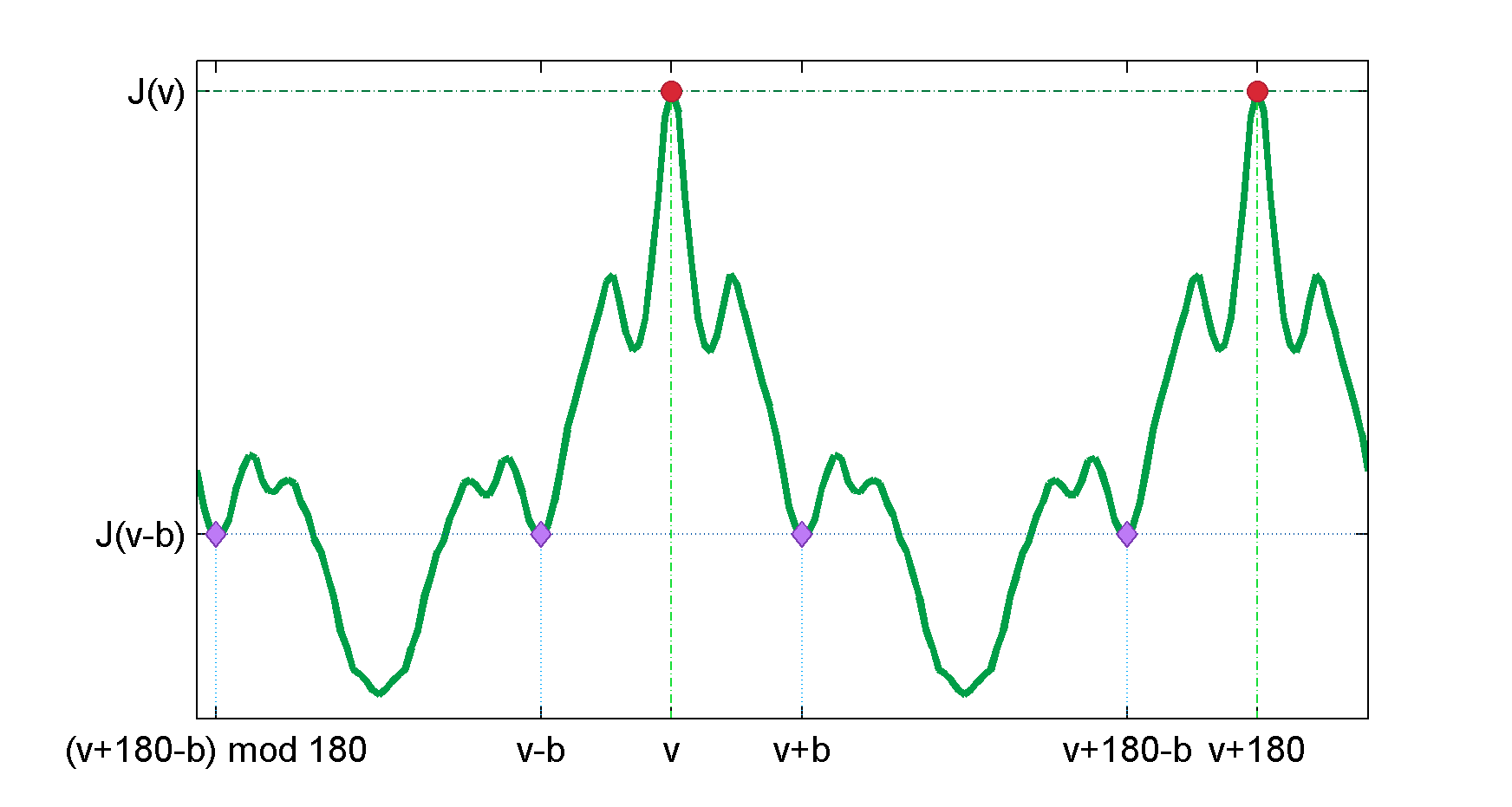}
\caption{\label{fig:symmetry}If an image has reflectional symmetry its reflectional negentropy posses a point reflection in $K$ points corresponding to the angles of image's reflection axes.}
\end{figure}

A graphical demonstration of Theorem~\ref{thm:theorem1} is given in \figref{fig:th1} and in \figref{fig:symmetry}. An immediate conclusion arising from Theorem~\ref{thm:theorem1} is presented in a lemma below. 

\begin{lemma}
\label{thm:lemma1}
Let $I$ be a reflectionally symmetric image with respect to a line $L$ rotated by $\theta$ about the origin, and let $\Delta$ be a distance between two neighbouring angles for which reflectional negentropy is calculated. 

Then the following holds.
\begin{enumerate}[label=(\alph*)]
\item If $J\left(\theta - \Delta\right) > J\left(\theta\right)$ then reflectional negentropy $J_{\Ref}\left(I, 
\cdot\right)$ attains local minimum at $\theta$. 
\item If $J\left(\theta - \Delta\right) < J\left(\theta\right)$ then reflectional negentropy $J_{\Ref}\left(I, \cdot \right)$ attains local maximum at $\theta$. 
\end{enumerate}
\end{lemma}

For clarity we will refer to the negentropy of the distribution of a reflectional negentropy $J\left(J_{\Ref}\right)$ as a \textit{point-symmetry negentropy}.

An outline of the tilt angle search algorithm is presented in Algorithm~\ref{alg:negTiltAngle}. For each local extremum the reflectional negentropy vector is shifted so that the extremum becomes the central element (we are allowed to do it, because reflectional negentropy is always a periodic function). A~reflected version of this vector is obtained by a left-right flip and finally the point-symmetry negentropy of an average of these two vectors is calculated. The extremum for which a relative error between its point-symmetry negentropy and the baseline negentropy is the smallest and no greater than $\delta$ is returned. If no eligible extremum is found then $-1$ is returned to indicate the absence of reflectional symmetry in the image.
 
\begin{algorithm}[htb]
\caption{\label{alg:negTiltAngle}NegTiltAngle}
\begin{algorithmic}[1]
\Procedure {NegTiltAngle}{$J_{\Ref}$, $extrema$}
\If {$\left\vert{extrema}\right\vert > 0$}
	\State $J \gets$ \Call{Negentropy}{$J_{\Ref}$}
	\State $errors \gets \varnothing$
	\For {$i = 1, \left\vert{extrema}\right\vert$ }
		\State $J_a \gets$ shift $J_{\Ref}$ by $(\left\vert{J_{\Ref}}\right\vert + 1)/2 - extrema[i]$
		\State $J_b \gets$ flip $J_a$ left-right
		\State $J_p \gets$ \Call{Negentropy}{$(J_a + J_b)/2$}
		\State $errors[i] \gets |J - J_p|/J$
	\EndFor
	\State $candidate \gets \argmin_i : errors[i]$
	\If {$errors[candidate] \leq \delta$}
		\State \Return $candidate$
	\EndIf
\EndIf
\State \Return -1
\EndProcedure
\end{algorithmic}
\end{algorithm}

\subsection{Combining All the Steps}

The \textsc{NegSymmetry} algorithm takes four input parameters: an input image $I$, an upper bound of the order of symmetry $K_{\max}$, an error threshold $\delta$, and a precision $\Delta$ used for spacing the angle domain when calculating reflectional negentropy. At first, the algorithm determines eligible orders of symmetry. Starting from the largest candidate order, the algorithm checks the reflectional negentropy for periodicity. Finally, the tilt angle is retrieved if it exists. 
The pseudocode of the algorithm is given in Algorithm~\ref{alg:negsymmetry}.

\begin{algorithm}[htb]
\caption{\label{alg:negsymmetry}NegSymmetry}
\begin{algorithmic}[1]
\Procedure {NegSymmetry}{$I$, $K_{\max}$, $\delta$, $\Delta$}
\State $result \gets \{order : 1, \theta_0 : -1\}$ \Comment{no symmetry}
\State $J_{\Rot} \gets$ \Call{RotationalNegentropy}{$I$, $K_{\max}$} 
\State $J_{\Ref} \gets$ \Call{ReflectionalNegentropy}{$I$, $\Delta$} 
\State $J_1 \gets J_{\Rot}[1]$
\State $orders \gets \{\,k : |J_{\Rot}\left(k\right) - J_1 | / J_1 \leq \delta \,\}$ 
\State $i \gets$ \Call{Length}{$orders$} 
\While {$i > 1$}
	\State $order \gets orders[i]$
	\If {\Call{IsPeriodic}{$J_{\Ref}$, $order$}} 
		\State $result.order \gets order$ 
		\State $extrema \gets \{\, k : |J_{\Ref}\left(k\right) - J_1|/J_0 \leq \delta \And J_{\Ref}\left(k\right) \textrm{ is a local extremum }\, \}$		
		\State $result.\theta \gets$ \Call{NegTiltAngle}{$J_{\Ref}$, $extrema$} 
		\State \Break
	\EndIf
	\State $i \gets i - 1$	
\EndWhile  
\State \Return $result$
\EndProcedure
\end{algorithmic}
\end{algorithm}

\section{\label{sec:experiments}Experimental Verification}

In order to verify that the proposed information theoretical principle for symmetry detection is valid an extensive experimental verification was conducted. The tests were executed on a set of $794$ symmetric images, $256$ by $256$ pixels large, with both reflectional and rotational types of symmetry, and with the order of symmetry ranging from $2$ to $9$ in case of rotational symmetry and from $1$ to $9$ in case of reflectional symmetry. During the preprocessing stage images were resized to a desired resolution and converted to greyscale.

\begin{figure*}
\centering
\experiment{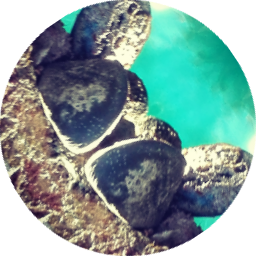}{exp1}
\caption{\label{fig:exp1}Reflectional symmetry of order $1$. \ref{fig:exp1a} -- An original image. \ref{fig:exp1b} -- The plot of rotational negentropy in a function of an order of symmetry. The true order of symmetry is $1$ therefore the rotational negentropy does not attain the baseline negentropy value for any other order.  \ref{fig:exp1c} -- A greyscale image with a detected reflection axis. \ref{fig:exp1d} -- The plot of reflectional negentropy in the function of an angle. Local extremum at an angle $146^\circ$ gives the tilt angle of a reflection axis shown in \ref{fig:exp1c}.}
\end{figure*}

\begin{figure*}
\centering
\experiment{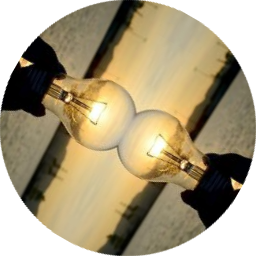}{exp2}	
\caption{\label{fig:exp2}Reflectional symmetry of order $1$. \ref{fig:exp2a} -- An original image. \ref{fig:exp2b} -- The plot of rotational negentropy in a function of an order of symmetry. \ref{fig:exp2c} -- A greyscale image with a detected reflection axes. \ref{fig:exp2d} -- The plot of reflectional negentropy in the function of an angle. Local extremum at an angle $120^\circ$ gives the tilt angle of a reflection axis shown in \ref{fig:exp2c}.}
\end{figure*}

\begin{figure*}
\centering
\experiment{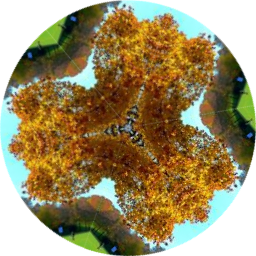}{exp3}
\caption{\label{fig:exp3}Reflectional symmetry of order $3$. \ref{fig:exp3a} -- An original image. \ref{fig:exp3b} -- The plot of rotational negentropy in the function of an order of symmetry. The baseline negentropy value is attained for an order $3$. \ref{fig:exp3c} -- A greyscale image with the detected reflection axes. \ref{fig:exp3d} -- The plot of reflectional negentropy in the function of an angle. Local extrema at $50^\circ$, $110^\circ$ and $170^\circ$ give the tilt angles of the reflection axes shown in \ref{fig:exp3c}.}
\end{figure*}

\begin{figure*}
\centering
\experiment{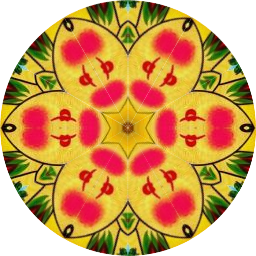}{exp5} 
\caption{\label{fig:exp5}Reflectional symmetry of order $6$. \ref{fig:exp5a} -- An original image. \ref{fig:exp5b} -- The plot of rotational negentropy in the function of an order of symmetry. The baseline negentropy value is attained for prime orders $2$ and $3$, and their product $6$. The maximal order for which the baseline negentropy value was attained is $6$ and so it is chosen as the true order of symmetry. \ref{fig:exp5c} -- A greyscale image with the detected reflection axes marked on top of it. \ref{fig:exp5d} -- The plot of reflectional negentropy in the function of an angle has 6 equidistant extrema, each of which is a centre of a point symmetry.}
\end{figure*}

\begin{figure*}
\centering
\experiment{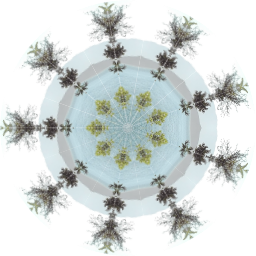}{exp6}
\caption{\label{fig:exp6}Reflectional symmetry of order $9$. \ref{fig:exp6a} -- An original image. \ref{fig:exp6b} -- The plot of rotational negentropy in the function of an order of symmetry. The baseline negentropy value is attained for prime order $3$ and its square $9$. As $9$ is the largest candidate found it is considered the detected order of symmetry. \ref{fig:exp6c} -- A greyscale image with the detected symmetry axes. \ref{fig:exp6d} -- The plot of reflectional negentropy in the function of an angle has nine local equidistant extrema. }
\end{figure*}

\begin{figure*}
\centering
\experiment{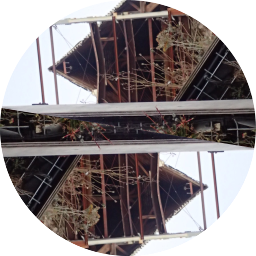}{exp7} 
\caption{\label{fig:exp7}Rotational symmetry of order $2$. \ref{fig:exp7a} -- An original image. \ref{fig:exp7b} -- The plot of rotational negentropy in the function of an order of symmetry. The baseline negentropy value is attained for an order $2$. \ref{fig:exp7c} -- A greyscale image. No reflection axes were detected therefore no axes are marked. \ref{fig:exp7d} -- The plot of reflectional negentropy in the function of an angle. The function is periodic but does not exhibit any point symmetry. Moreover, no value falls into the eligible interval surrounding the baseline negentropy value. }
\end{figure*}

\begin{figure*}
\centering
\experiment{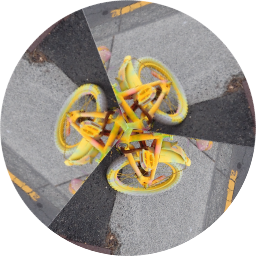}{exp8}	
\caption{\label{fig:exp8}Rotational symmetry of order $3$. \ref{fig:exp8a} -- An original image. \ref{fig:exp8b} -- The plot of rotational negentropy in the function of an order of symmetry. The baseline negentropy value  is attained for an order $3$. \ref{fig:exp8c} -- A greyscale image. As no reflection axes were found none is marked. \ref{fig:exp8d} -- The plot of reflectional negentropy in the function of an angle. The periodic function does not exhibit any point symmetry. Moreover, there is no local maximum within the eligible interval surrounding the baseline negentropy value. }
\end{figure*}

\begin{figure*}
\centering
\experiment{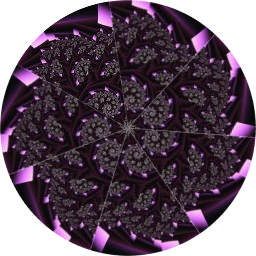}{exp9} 
\caption{\label{fig:exp9}Rotational symmetry of order $7$. \ref{fig:exp9a} -- An original image. \ref{fig:exp9b} -- The plot of rotational negentropy in the function of an order of symmetry. The baseline negentropy value is attained for an order $7$. \ref{fig:exp9c} -- A greyscale image. \ref{fig:exp9d} -- The plot of reflectional negentropy in the function of an angle. No local extremum that would meet the requirements was found.}
\end{figure*}

\begin{figure*}
\centering
\experiment{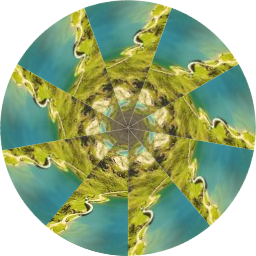}{exp10}	
\caption{\label{fig:exp10}Rotational symmetry of order $9$. \ref{fig:exp10a} -- An original image. \ref{fig:exp10b} -- The plot of rotational negentropy in the function of an order of symmetry. The baseline negentropy value is attained for an order $9$ as well as for its prime factor $3$. \ref{fig:exp10c} -- A greyscale image. \ref{fig:exp10d} -- The plot of reflectional negentropy in the function of an angle. No local extremum that would meet the requirements was found. }
\end{figure*}

The examples of the accurately detected reflectional and rotational symmetries are presented in \crefrange{fig:exp1}{fig:exp10}. In each figure there are four images given. 

The top left subfigure (a) presents an original image prior to its conversion to greyscale. 

The top right subfigure (b) presents the plot of rotational negentropy as a function of the order of rotational symmetry, as calculated for a greyscale image. 
The dashed lines indicate tolerance bounds on the image's baseline negentropy, as given in Equation~\eqref{eq:toleranceBounds}. When analysing this plot one is interested in finding the arguments for which rotational negentropy falls within the bounds, as these arguments correspond to the orders of symmetry exhibited by an image. Consequently, in \cref{fig:exp1,fig:exp2} the detected order of symmetry is $1$ as the value of baseline negentropy is not attained for any other argument, while for \crefrange{fig:exp3}{fig:exp10} higher orders of symmetries are detected. Another noteworthy characteristic of an order of symmetry is that if it is a prime number then the value of the baseline negentropy is attained only for this particular order. For instance, in \cref{fig:exp3b,fig:exp8b} the baseline negentropy values are attained only for an order $3$, in \cref{fig:exp7b} only for an order $2$ and in \cref{fig:exp9b} only for an order $7$. If, on the other hand, an order of symmetry is not prime then the value of the baseline negentropy is attained for this order as well as for its prime factors and their products. For instance in \cref{fig:exp5b} the baseline negentropy value is attained for the order $6$ as well as for $2$ and $3$ while in \cref{fig:exp6b,fig:exp10b} for the order $9$ as well as for its factor $3$. 

The bottom left subfigure (c) presents an input image converted to greyscale and with the detected reflection axes marked on top of it, if any reflection axes were found. 

The bottom right subfigure (d) is a plot of reflectional negentropy as a function of the angle of the reflection axis.
The dashed lines indicate baseline negentropy bounds, which are exactly at the same levels as in (b). 
Given the plot of reflectional negentropy one searches for such local extrema that fall within these bounds and, at the same time, are centres of point reflections. 
The arguments for which eligible extrema are found are the detected angles of reflection axes. Consequently, the detected type of symmetry is reflectional. 
However, if no eligible extrema are found then the detected type of symmetry is rotational. 
Among examples given below, \crefrange{fig:exp1}{fig:exp6} are the examples of reflectional symmetry. Eligible extrema are marked with circular green point markers (d) and the reflection axes are plotted on top of a greyscale image (c). \crefrange{fig:exp7}{fig:exp10} are the examples of rotational symmetry. Local extrema of reflectional negentropy are either below the lower bound (\cref{fig:exp7,fig:exp9}) or above the upper bound (\cref{fig:exp8,fig:exp10}). 





\subsection{Failed detection}
\begin{figure*}
\centering
\experiment{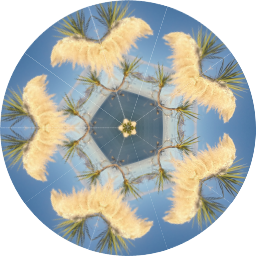}{failed1}	
\caption{\label{fig:failed1}Reflectional symmetry of order $5$. \ref{fig:failed1d} -- The other local extremum within an acceptance interval was found to give better point symmetry and as such it was incorrectly selected as the one corresponding to the true angle of symmetry axis. }
\end{figure*}

In \figref{fig:failed1} an example of a failed detection is presented. In this example an order of symmetry is correctly recognised to be $5$ but the angles of reflection axes are missed by $18^\circ$. The plot \figref{fig:failed1d} reveals that there are two sets of local extrema - at $k \cdot 180^\circ / 5$ and at $18 + k \cdot 180^\circ / 5$, $k=0,\dots,4$. The former set is correct, but the latter is chosen by the detector due to the less erroneous point symmetry. Images for which reflectional negentropy has eligible extrema not associated with reflection axes pose a challenge to the detector. 

\subsection{Comparison of the Symmetry Detection Methods}
Following the method validation step reported above we compared our method to two other existing methods. 
\begin{enumerate}
\item A local symmetry detector proposed by Loy and Eklundh~\cite{loy2006detecting} which was reported to outperform other algorithms in two recent competitions of symmetry detectors~\cite{rauschert2011first, liu2013symmetry}. Its implementation is publicly available~\cite{loy2006code} and was used in our experiments. 
\item The Shen-Ip Symmetries Detector \cite{shen1999symmetry}, one example of a class of approaches that uses moments and moment invariants in symmetry detection. This method was selected due to its conciseness as well as due to recent development of its extension to non-centred symmetries~\cite{tzimiropoulos2009unifying}.
\end{enumerate}

In what follows we compare methods' success rates for three tasks: detection of the angle of a reflection axis if such exists, detection of the order of symmetry and detection of the type of symmetry in images. We used the same image set as in the previous experiment. Additional implementation and assessment details for both methods are given in Supplementary material available on our website~\cite{migalska2016code}.

In Table~\ref{tab:angleDetection} the success rates for the detection of the angles of reflection axes are given. There were $436$ reflectionally symmetric images in the test dataset. We assess the angle detection by three criteria -- exact, strict and lenient. The criteria are based on the comparison of the retrieved angle with the ground truth.
The exact criterion is satisfied if the two angles are the same. The strict and lenient criteria are satisfied if the absolute difference between the retrieved angle and the ground truth does not exceed $2^\circ$ and $10^\circ$, respectively. From the results it is visible that our method, \textsc{NegSymmetry}, has the highest success rate in accordance with the exact criterion and performs slightly worse than the Loy \& Eklundh's algorithm in accordance with the strict and lenient criteria. Our method outperforms the Shen-Ip detector regardless of the criterion used.

\begin{table}[H]
\centering
\begin{tabular}{crrrrr}\toprule
\rot{\shortstack[l]{Algorithm}} & \multicolumn{1}{c}{\rot{\shortstack[l]{Image\\ Resolution}}} & \multicolumn{1}{c}{\rot{\shortstack[l]{Relative\\Error $\delta$}}} & \multicolumn{1}{c}{\rot{\shortstack[l]{Exact\\Criterion}}} & \multicolumn{1}{c}{\rot{\shortstack[l]{Strict\\Criterion}}} & \multicolumn{1}{c}{\rot{\shortstack[l]{Lenient\\Criterion}}} \\\midrule
NS & $128$ & $0.05$ & $87.61\%$ & $88.53 \%$ & $90.60\%$  \\
NS & $128$ & $0.1$ & $87.61\%$ & $88.30 \%$ & $90.37\%$ \\ 
LE & $128$ & N/A & $31.65 \%$ & $98.61 \%$ & $99.70 \%$  \\
SI & $128$ & N/A & $46.79 \%$ & $55.73 \%$ & $60.09 \%$ \\\midrule
NS & $256$ & $0.05$ & $89.68 \%$ & $91.06\%$ & $92.66 \%$ \\
NS & $256$ & $0.1$ & $87.61 \%$ & $88.99 \%$ & $90.37\%$ \\
LE & $256$ & N/A & $29.82 \%$ & $99.77 \%$ & $100.00 \%$ \\
SI & $256$ & N/A & $42.43 \%$ & $47.48 \%$ & $52.29 \%$ \\
\bottomrule
\end{tabular}
\caption{\label{tab:angleDetection}Detection success rate of the angle of a reflection axis on a set of $436$  symmetric images with reflectional symmetry of order from $1$ to $9$ obtained by various symmetry detection algorithms: NS -- NegSymmetry, LE -- Loy and Eklundh, SI -- Shen-Ip. Success rate calculated based on three different criteria for the acceptable absolute difference between the ground truth angle and the detected angle --  exact (no difference), strict (no more than $2^\circ$) and lenient (no more than $10^\circ$s).}
\end{table}

In Table~\ref{tab:results} the orders of symmetry and success rates calculated on a set of $794$ images are presented. The proposed method clearly outperforms the other contestants in terms of the detection of both the order and the type of symmetry. The results also demonstrate that the relative error $\delta$ should be chosen with diligence, as the larger error tolerance gives higher success rate on a symmetry order detection while a more strict approach results in a higher success rate on the symmetry type detection. 

\begin{table}[H]
\centering
\begin{tabular}{crrrrrr}\toprule
\rot{\shortstack[l]{Algorithm}} & 
\multicolumn{1}{c}{\rot{\shortstack[l]{Image\\ Resolution}}} & 
\multicolumn{1}{c}{\rot{\shortstack[l]{Relative\\Error $\delta$}}} & 
\multicolumn{1}{c}{\rot{\shortstack[l]{Symmetry Order\\Detection Rate}}} &
\multicolumn{1}{c}{\rot{\shortstack[l]{Symmetry Type\\Detection Rate}}} \\\midrule
NS & $128$ & $0.05$ & $85.89\%$ & $87.53\%$ \\
NS & $128$ & $0.1$ & $91.06\%$ & $81.74\%$ \\
LE & $128$ & N/A &  $81.84 \%$ & $58.64 \%$ \\
SI & $128$ & N/A & $60.71 \%$ & $59.45 \%$ \\\midrule
NS & $256$ & $0.05$ & $88.29\%$ & $89.29\%$ \\
NS & $256$ & $0.1$ & $93.07\%$ & $84.51\%$ \\
LE & $256$ & N/A & $80.48 \%$ & $54.91 \%$ \\
SI & $256$ & N/A & $18.89 \%$ & $49.12 \%$ \\\bottomrule
\end{tabular}
\caption{\label{tab:results}Success rates for the detection of the order of symmetry and the type of symmetry on a set of $794$ symmetric images with either reflectional symmetry of order $1$ to $9$ or rotational symmetry of order $2$ to $9$. Compared methods: NS -- NegSymmetry, LE -- Loy and Eklundh, SI -- Shen-Ip. }
\end{table}

\section{\label{sec:control}Application to Automatic Visual Inspection}  

Modern manufacturing is increasingly characterized by an in-process inspection to control production and achieve the desired quality, rather than by an acceptance or rejection at the end. In most manufacturing industries, the goal is to achieve 100\% quality assurance of the parts, sub-assemblies, and finished products. Visual inspection in production lines has become an important step in the manufacture process \cite{zhao2009computer}, which is not only aimed at guaranteeing that the manufactured goods are of the highest quality but also at reducing production time and cost. Inspection is the process of determining if a product (a part, object, item) deviates from a given set of specifications \cite{newman1995survey}. Inspection usually involves measurement of specific part features such as assembly integrity, surface finish and geometric dimensions. Among these checks, shape inspection is one of the key inspections in industry \cite{vernon1991machine, malamas2003survey}.

An inspection starts with a camera that scans an item and submits its image to machine vision systems for the compliance with the standards verification, for instance with the requirements for shape. In many instances this shape requirements can be expressed in terms of symmetries, for instance the alloy wheels might be required to exhibit reflectional symmetry of order 5 while a reflectional symmetry of order 3 might be mandatory in a planetary gear system. The absence of symmetry in these cases indicates that the geometry of a product is distorted, and so the product does not adhere to requirements and may pose a significant threat to its users. 

\begin{figure*}
\centering
\experiment{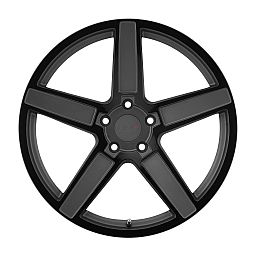}{alloy_wheels_1} \\
\caption{\label{fig:alloy_wheels_1}Frontal view of alloy wheels. An expected reflectional symmetry of order $5$ is detected by the method. In \ref{fig:alloy_wheels_2} the shape of the wheel is distorted and the symmetry is no longer present.}
\end{figure*}

\begin{figure*}
\centering
\experiment{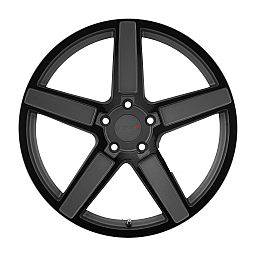}{alloy_wheels_2}
\caption{\label{fig:alloy_wheels_2}Frontal view of alloy wheels. The shape of the wheel is distorted and the symmetry is no longer present.}
\end{figure*}

Below an example of a product geometry verification is given. An alloy wheel in \figref{fig:alloy_wheels_1} exhibits required symmetries and so it passes visual inspection. On the other hand, a distorted wheel in \figref{fig:alloy_wheels_2} does not exhibit the desired symmetries. 

\section{\label{sec:conclusion}Discussion and Conclusion}

Motivated by demands arising from quality control that existing methods for symmetry detection failed to satisfy we have proposed a novel method for symmetry detection based on information theoretical principles. The detection of a symmetry transformation within an image is achieved by comparing image's baseline negentropy, that is the negentropy of an original image, to the negentropy of an average of intensities of the original image and of its transformed copy. If the symmetry transformation is correctly chosen, an absolute difference between the negentropy of an averaged image and the baseline negentropy does not exceed a preselected relative error $\delta$. Otherwise, the absolute difference exceeds $\delta$ and, relative to this particular accuracy level, the transformation cannot be considered a proper symmetry transformation.

In the proposed method we have utilized the measures rooted in information theory to assess the changes in the amount of information transmitted by a message. The employment of entropy instead of other statistics is justified by the fact that it is a fundamental and compact characterization of a probability distribution. The expansion of entropy into a Fourier series is an infinite weighted sum of moments, thus entropy reflects all moments of the distribution. Further, the proposed detector is ``negentropic'' due to the fact that negentropy calculation is both sufficient (as explained in section~\ref{sec:theory}) and also computationally simpler than entropy computation using the adopted approximation \eqref{eq:entropyApproximation}. 

Moreover, we have shown that expressing the problem of symmetry detection in terms of its negentropy allows reducing the dimensionality of the problem. The problem of detecting the order of symmetry in a  two-dimensional image can be simplified to a one-dimensional problem of periodicity identification in the image's reflectional negentropy vector, while the problem of detecting the angle of a symmetry axis can be simplified to a one-dimensional problem of finding a point-symmetry in the same vector.

An extensive experimental verification of our algorithm conducted on a large set of symmetric images, with the symmetry order, type and angle of a symmetry axis varying among the examples, allowed us to conclude that the proposed detector is superior to other methods on the task of global symmetry detection in images. While we admit that the set of compared methods is limited, we make our set of images along with its symmetry ground truth publicly available. This benchmark set can be successfully employed to assess the performance of other symmetry detection methods. 

Finally, we have demonstrated that our method can be successfully applied to assure the required quality of manufactured goods. An example of a deformed alloy rim demonstrates that symmetry detection can be a suitable tool to solve problems in quality control. 

Future work encompasses a reformulation of the proposed method into the framework of statistical tests for symmetry.  

\section*{Acknowledgements}
The work of Agata Migalska has been supported by the National Science Center under grant: 2012/07/B/ST7/01216, internal code 350914 of the Wroc\l{}aw University of Technology.

\bibliographystyle{elsarticle-num}
\bibliography{refs}

\appendix
\numberwithin{equation}{section}
\section{\label{sec:appendix1}Proof of Property~\ref{thm:property1}}

Let $K$ be an order of reflectional symmetry.  For $K=1$ the existence of simultaneous rotational symmetry of an order $1$ follows directly from Property~\ref{thm:property1}. For $K\geq 2$ a proof is based on the fact that combining any two reflections yields a rotation. Precisely, a combination of two reflections in distinct intersecting lines is a rotation about the point of intersection (here: the origin and also the centre of an image) by twice the angle between the two mirror lines.
\begin{align}
\begin{split}
\left(\Ref_\phi \cdot \Ref_\theta\right)\left(x,y\right) &= \left(\begin{array}{lr}\cos 2\phi & \sin 2\phi \\ \sin 2\phi & -\cos 2 \phi \\\end{array}\right) \left(\begin{array}{lr} \cos 2\theta & \sin 2\theta \\ \sin 2\theta & -\cos 2\theta \end{array}\right) \left( \begin{array}{c}x \\ y\end{array}\right) = \\
&= \left(\begin{array}{lr} \cos 2\left(\phi - \theta\right) & -\sin 2\left(\phi - \theta\right) \\ \sin 2\left(\phi - \theta\right) & \cos 2\left(\phi - \theta\right) \end{array}\right) 
\cdot \left( \begin{array}{c}x \\ y\end{array}\right) = \Rot_{2\left(\phi - \theta\right)}\left(x,y\right)
\end{split}
\label{eq:2reflEqRot}
\end{align}

An image has a reflectional symmetry of order $K$ if it is invariant to reflection with respect to the reflection axes rotated by $\theta_0 + \frac{k\cdot 180^\circ}{K}$, $k=0,\dots, K-1$, about the origin. An image has a rotational symmetry of order $K$ if it is invariant to rotation by $\frac{l\cdot 360^\circ}{K}$, $l=0,\dots, K-1$, about the origin. It is enough to demonstrate that all the above $K$ rotations and none other rotation are achieved by a combination of two reflections.

Let $k_1, k_2, k_3 \equiv \{0,\dots, K-1\} \mod K$. Let $\Ref\left(\theta_0 + \frac{k_1\cdot 180^\circ}{K}\right)$ and $\Ref\left(\theta_0 + \frac{k_2\cdot 180^\circ}{K}\right)$ be two arbitrary reflections. Then from \eqref{eq:2reflEqRot} it immediately follows that

\begin{proof}
\begin{align*}
\begin{split}
& \Ref\left(\theta_0 + \frac{k_1\cdot 180^\circ}{K}\right) \cdot \Ref\left(\theta_0 + \frac{k_2\cdot 180^\circ}{K}\right) = \Rot\left(2\left(\theta_0 + \frac{k_1\cdot 180^\circ}{K} - \theta_0 - \frac{k_2\cdot 180^\circ}{K}\right)\right) = \\
& = \Rot\left(\frac{\left(k_1 - k_2\right)\cdot 360^\circ}{K}\right) = \Rot\left(\frac{k_3\cdot 360^\circ}{K}\right)
\end{split}
\end{align*}
\end{proof}

\end{document}